%% file: Main.tex
\documentclass[runningheads]{llncs}
\usepackage{graphicx}
\usepackage{caption}
\usepackage{subcaption}
\usepackage[table]{xcolor} 
\usepackage{sparklines}
\usepackage{hyperref}
\usepackage[misc,geometry]{ifsym}
\begin{document}
\title{SpaDen : Sparse and Dense Keypoint Estimation for Real-World Chart Understanding}
\titlerunning{SpaDen: Sparse-Dense Chart Keypoint Estimation}
%
\author{Saleem Ahmed$^\dagger$(\Letter)\orcidID{0000-0001-8648-9625} ,
Pengyu Yan$^\dagger$\orcidID{0000-0003-1584-2350}, 
David Doermann\orcidID{0000-0003-1639-4561},
Srirangaraj Setlur\orcidID{0000-0002-7118-9280}, 
Venu Govindaraju\orcidID{0000-0002-5318-7409}
}

\authorrunning{S.Ahmed et al.}

\institute{Department of Computer Science and Engineering,\\
University at Buffalo, SUNY \\
\email{\{sahmed9,pyan4,doermann,setlur,govind\}@buffalo.edu}
}

%
%

%
\maketitle              
\begin{abstract}

We introduce a novel bottom-up approach for the extraction of chart data. Our model utilizes images of charts as inputs and learns to detect keypoints (KP), which are used to reconstruct the components within the plot area. Our novelty lies in detecting a fusion of continuous and discrete KP as predicted heatmaps. A combination of sparse and dense per-pixel objectives coupled with a uni-modal self-attention-based feature-fusion layer is applied to learn KP embeddings. Further leveraging deep metric learning for unsupervised clustering, allows us to segment the chart plot area into various objects. By further matching the chart components to the legend, we are able to obtain the data series names. A post-processing threshold is applied to the KP embeddings to refine the object reconstructions and improve accuracy. Our extensive experiments include an evaluation of different modules for KP estimation and the combination of deep layer aggregation and corner pooling approaches. The results of our experiments provide extensive evaluation for the task of real-world chart data extraction. 
\footnote{This is a pre-print version. Accepted at ICDAR'23}.

\end{abstract}
\input{Sections/001_Introduction.tex}
\input{Sections/002_Background.tex}
\input{Sections/003_Methodology.tex}

\input{Sections/004_Results.tex}

\bibliographystyle{splncs04}
\bibliography{Main}
\end{document}

%% file: Sections/001_Introduction.tex
\section{Introduction}
Data visualizations are effective constructs to efficiently convey knowledge in documents. Most documents consist of semantically structured textual content and complementary visualizations in the form of figures, generic infographics, technical diagrams, charts, etc. Our work focuses on the problem of reconstructing tabular data used to plot the visualization, given the image and structural information as input. We focus on the family of visualizations called charts, specifically from the scientific literature.

\subsection{Charts}
Charts are visual representations of data composed of simple abstract shapes arranged to have a semantic meaning. They are often used to make it easier to understand large amounts of data and to see patterns and trends. Charts can represent many types of data, including numerical, categorical, and time-based data. Standard chart types include bar, line, scatter, and box plots. The elements of a chart are the visual components used to represent data. These can include the title of the chart, plot text labels, the axes, the legend, and the data points or lines.

The chart's title is typically a brief phrase or sentence describing the data being plotted. The plot text labels include text such as the actual values represented on the chart. The axes are the lines along the bottom/top and left/right sides of the chart and show the scale and range of the data being plotted. The axis usually has an axis title, major/minor tick marks, and tick labels. The legend is a key that explains the meaning of different colors, symbols, or other visual elements used in the chart. These are usually in a horizontal or vertical box with patches representing the key and text representing a data series. The data points or lines are the individual elements of the chart that represent the data. For example, in a line chart, the data points would be plotted along the line, while in a bar chart, they would be represented by the individual bars, grouped bars, or stacked bars; box plots show the distribution as a box with three whiskers - first quartile, central tendency, and third quartile, most box plots also have two whiskers for minimum and maximum value whereas, in a scatter plot, they are just the points.

Other chart elements can include gridlines, which are lines that help to divide the chart into smaller sections and make it easier to read, and data labels, which show the exact values for each data point or line. The overall design of the chart, including the colors, fonts, and layout, can also be considered an element of the chart.

\subsection{Chart Data Extraction} 

There has been decent strides made in the space of document understanding \cite{appalaraju2021docformer}, \cite{gu2021unidoc}, \cite{zhong2019publaynet}, \cite{xu2020layoutlm}, with specialised tasks for non-textual understanding such as parsing reason over mathematical expressions \cite{9412619},\cite{mansouri2022advancing} One such focus has been automation of chart parsing, which originated from public challenges for chart data extraction\cite{icpr2019,icpr2020,icpr2022}. Multiple iterations of this competition have spurred significant community interest in this highly challenging task. While earlier challenges comprised large-scale (~100k images) synthetic charts on which deep models can achieve very high accuracies, the latest iterations feature only real-world charts, which continue to be extremely challenging especially as it pertains to the end-to-end data extraction task.  
Our focus is on Task-6 of the challenge where the input is a chart image, the text corresponding to the chart, and structural properties such as the role of each text element and legend and axes elements, and the output is a table with the data used to generate the original chart image.  
Other works outside this competition have also been published on the chart data extraction task but remain severely constrained, as discussed in section \ref{sec:domain}.   


%% file: Sections/002_Background.tex
\section{Background}

In this section, We describe our problem , popular model architectures used in the literature for keypoint(KP) and object detection tasks and techniques for improving such models.
We also discuss relevant prior works in this domain. 
\subsection{Chart Infographics : Chart Data Extraction Challenge}
This challenge \cite{icpr2022} aims to evaluate and promote the development of automated chart data processing systems. This involves the extraction of structured data from chart images. The challenge is divided into six sub-tasks, which mimic the common steps used in manual chart data extraction.

The first task is chart type classification. The second task is the detection and recognition of text regions in the input chart image. In the third task, text elements are classified according to their semantic role in the chart, such as chart title, axis title, or tick label. The fourth task requires associating tick labels with specific pixel coordinates. The fifth task involves pairing the textual labels in the legend with the associated graphical markers in the chart. 

The sixth task is data extraction, where the goal is to extract the original data used to create the chart. This is further divided into two parts: plot element detection and classification, and data extraction. The former involves segmenting the chart image into atomic elements such as bars, points, and lines, while the latter involves producing named sequences of $(x,y)$ pairs that represent the data points used to create the chart. Subsequent tasks assume output of previous tasks as available input.
For our work, we focus on the sixth task of this challenge.

 \subsection{Keypoint Estimation Architectures}

Keypoint estimation is a common task in computer vision, involving the detection of specific points or landmarks in an image. Popular models for this task include Hourglass Network(HGN), \cite{newell2016stacked}, Cascade Pyramid Network (CPN) \cite{chen2018cascaded}, and Simple Pose Network (SPN)\cite{li2020simple}. HGN has a bottleneck shape that compresses and expands data through downsampling and upsampling layers. Its symmetrical shape allows the model to learn spatial information at multiple scales. CPN is a variant of HGN that combines the strengths of bottom-up and top-down approaches using multiple cascaded hourglass modules. SPN, on the other hand, is a simpler and more efficient model than HGN and CPN, using devonvolution learnable layers for upsampling and a bottom-up approach to KP estimation. In the realm of KP localization benchmarks, including tasks such as human pose, facial landmark detection, and document corner detection, it is generally observed that while HGN and CPN are more accurate, they are also more complex and resource-intensive. Conversely, SPN is faster and simpler to train, but may sacrifice some degree of accuracy.

We conduct experiments with all three variants to provide an exhaustive comparison.

 \subsection{Anchor-free Object Detection for Keypoints}
 In the domain of object detection, a promising approach is to treat it as a KP detection problem and bypass the need for predefined anchors or bounding boxes. This anchor-free paradigm has been successfully implemented in the Single Shot Object Detection framework. CornerNet\cite{law2018cornernet} is based on an HGN backbone and uses two heads to detect the top-left and bottom-right KP of an object. The KP are combined using corner pooling layers. CentreNet\cite{duan2019centernet}extends CornerNet by using an additional center KP. Thus three KP are combined using centre pooling layers. Object as Points\cite{objaspt} proposes that the center point is sufficient to detect objects. It uses the expected center of a box as both an object and a KP to determine the coordinates and offsets of the bounding box. They combine DLA-based architectures \cite{yu2018deep}, with deformable convolutions replacing the upsampling layers in SPN.

 Inspired by these methods, we propose an extension that combines all four top-down, bottom-up, left and right directional pooling for improved KP features. Furthermore, we modify the downsampling convolution layers in popular anchor-free object detection models such as HGN, CPN, and SPN with a DLA-34 architecture to experiment with alternative backbones. To achieve this, we implement a hierarchical deep-aggregation strategy for each of the downsampling layers in a KP-backbone encoder, which we believe will enhance the overall performance of the model.

\subsection{Chart Data Extraction Models}
\label{sec:domain}
Domain-specific models have been proposed for bottom-up chart data extraction techniques. In \cite{ma2021towards}, the authors propose an ensemble of different popular models for each task of box detection, point detection, and legend matching. For bar plots and box-plots, they use a Feature Pyramid Network with a ResNet backbone, and for points, they use a Fully Convolutional Network for producing heatmaps. Further, they train a separate feature extractor with triplet loss over legend patches for the legend linking task. In the same 2020 challenge, another submission \cite{luo2021benchmark} uses a CentreNet model for bar plots and a CentreNet with DLA-34\cite{yu2018deep} connections for box, line, and scatter plots having a different number of final layers per chart type. They do legend matching through HOG features. 

Authors of \cite{luo2021chartocr} propose using a CornerNet model with an added head for chart-type classification. Current literature \cite{icpr2022} provides a general framework for bottom-up chart data extraction. They primarily utilize off-the-shelf computer vision models that are either disconnected and trained for separate tasks, have different architectures for  different chart types, or solve only half the task of visual element detection and no legend matching. Extracting data from a chart without its contextual legend information is ineffective.  Also, since these works were benchmarked on older versions of datasets or datasets with missing chart types, it is hard to compare across methods.

We provide a systematic study that encompasses this family of architectures and techniques and propose the first model, to our knowledge, for complete chart data extraction.


%

\subsection{Contrastive Loss for Visual Element Reconstruction} For plot elements such as lines, the predicted KP need to be clustered or `reconstructed' in a bottom-up fashion. To train these clustering embeddings we experiment with two types of contrastive losses.
\subsubsection{Push-pull loss} 


operates by comparing the distances between the reference points and the actual data points in the embedding space. For data points similar to their reference point (i.e.,  belonging to the same class), the push-pull loss function will try to minimize the distance between them while  maximizing for the rest. 




Mathematically, the push-pull loss for KP detection can be defined as:

$$L =  (1-Y) * (max(0, d_p - d_n + m))^2 + Y * (max(0, d_n - d_p + m))^2$$

where L is the loss, $Y$ is a binary label indicating whether the inputs are similar $(Y=1)$ or dissimilar $(Y=0), d_p$ is the distance between the KP detected for the positive input, $d_n$ is the distance between the KP detected for the negative input, and $m$ is a margin hyperparameter that determines how far apart the KP should be.



\subsubsection{ Multi-Similarity Loss (MS)} is designed to encourage learning deep features that are discriminative between classes and similar within each class. 

The discriminative term of the loss function is given by:

$$L_{dis} = -\sum_{i=1}^{N} \log \frac{e^{f_{y_i}}}{\sum_{j=1}^{C} e^{f_j}}$$

where $N$ is the number of training examples, $C$ is the number of classes, $y_i$ is the class label of the $i$th training example, and $f_j$ is the predicted class score for the $j$th class. 

The similarity term of the loss function is given by:

$$L_{sim} = \frac{1}{2N} \sum_{i=1}^{N} \sum_{j=1}^{N} [y_i = y_j] \left( 1 - \frac{f_{y_i} - f_{y_j}}{\max(0, f_{y_i} - f_{y_j}) + \alpha} \right)$$

where $\alpha$ is a hyperparameter that controls the strength of the similarity regularization. 

The sum of the discrimination and similarity terms then gives the overall loss function:

$$L = L_{dis} + L_{sim}$$

\input{Sections/Figures/00001archi}

%% file: Sections/Figures/00001archi.tex
\begin{figure}
     \centering
     \begin{subfigure}[b]{\textwidth}
         \centering
         \includegraphics[width=\textwidth]{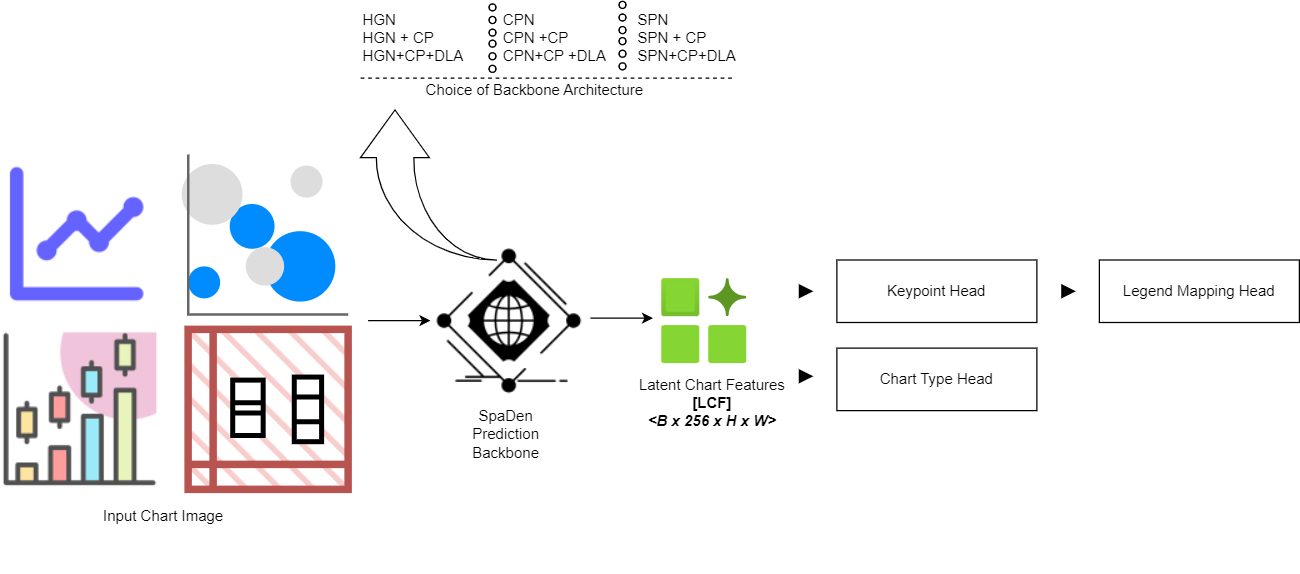}
         \caption{\scriptsize{The Unified Data Extraction(UDE) Framework}}
         \label{fig:arch-ude}
     \end{subfigure}
     \hfill
     \hfill
     \\
     \begin{subfigure}[b]{0.6\textwidth}
         \centering
         \includegraphics[width=\textwidth]{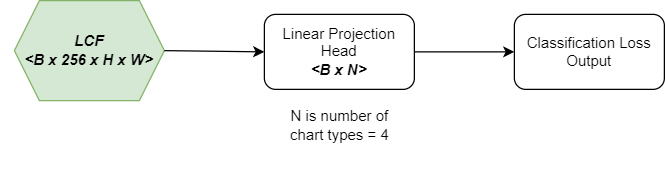}
         \caption{\scriptsize{Chart Type Head }}
         \label{fig:arch-cls}
     \end{subfigure}
     \hfill
     \hfill
     \\
     \begin{subfigure}[b]{1.0\textwidth}
         \centering
         \includegraphics[width=\textwidth]{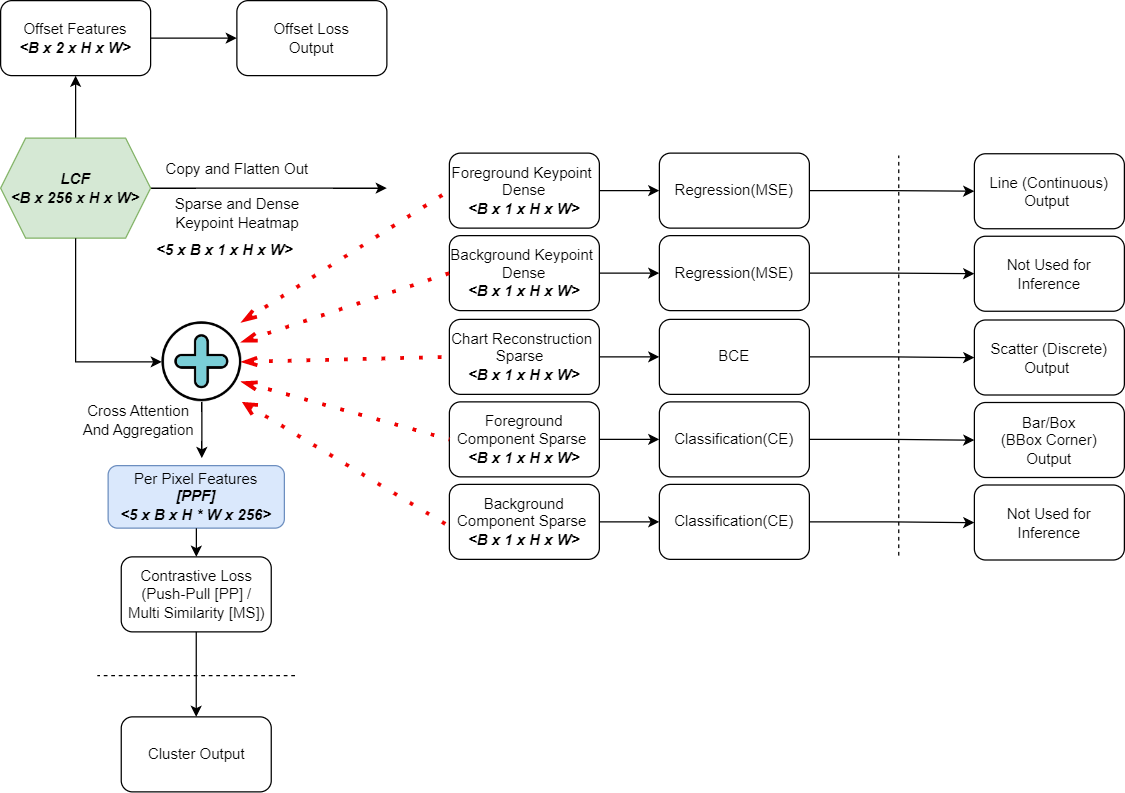}
         \caption{\scriptsize{SpaDen Keypoint Head : All 5 heatmaps are trained for each chart type. During inference selective output is used. The dense/sparseness comes from the segmentaion mask used for training.}}
         \label{fig:arch-kp}
     \end{subfigure}
     \hfill
     \hfill
     \\
     \begin{subfigure}[b]{0.8\textwidth}
         \centering
         \includegraphics[width=\textwidth]{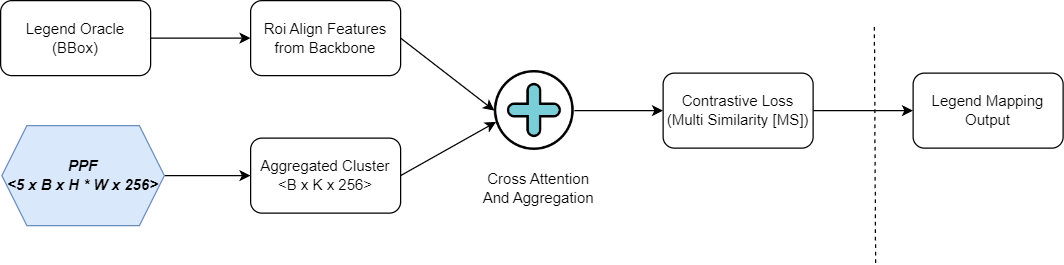}
         \caption{\scriptsize{Legend Mapping Head}}
         \label{fig:arch-llm}
     \end{subfigure}
     \hfill
        \caption{Chart Data Extraction using SpaDen Model in the UDE Framework.}
        \label{fig:UDE}
\end{figure}

%% file: Sections/003_Methodology.tex
\section{Unified Data Extraction (UDE) Framework}
\label{sec:UDE}
We propose a generic framework for bottom-up parsing for chart data extraction. Fig. \ref{fig:ude} shows the general blocks of UDE and Fig.\ref{fig:UDE} shows a more detailed view of the overall architecture. The UDE framework consists of a  generalized backbone chart feature extractor, chart type prediction, KP localization, chart component reconstruction by KP grouping, and a legend mapping block. We discuss these blocks in detail in this section. We also describe a custom segmentation mask formulation, and making the feature extractor invariant to chart text.   
\input{Sections/Figures/002_UDE.tex}
\subsection{Backbone Feature Extractor} 
Inspired by architectures used for KP extraction, our baselines consist of HGN, CPN, SPN, DLA-34, and general-purpose Resnet. 
We further propose a novel architecture using DLA techniques. We modify the connection layers of HGN, CPN, and SPN by adding iterative deep aggregations. Another way of understanding these is (i) Hourglass blocks where each stage is a DLA encoder, (ii) DLA model with hybrid top-down and bottom-up output layers similar to CPN and (iii) DLA model with deconvolution layers instead of upsampling as in SPN.
We conduct ablation studies with each variant and corner pooling layers.

\subsection{Chart Type Classification} 
A linear projection head takes base features and passes them through a series of 1x1Conv-BN layers to get the chart-type output. This is shown in Fig. \ref{fig:arch-cls}

\subsection{Keypoint Localization}
These networks are trained with an L2  custom loss for KP and an L1 distance loss for offset.We refer readers to \cite{law2018cornernet} for their detailed implementations. Generally, KP estimation literature makes an architectural decision based on the number of output classes for this head. In charts, we have an unconstrained number of KP for types like scatter and line. In our implementation, we predict a single heatmap for all KP. As depicted in Fig. \ref{fig:arch-kp} we predict 5 different `views' of the KP heatmap, trained using different segmentation masks discussed in detail below.

\subsubsection{Dense Directional Keypoint Masks} \input{Sections/Figures/004_masks}
are developed to learn regression heatmaps for unconstrained KP in charts. This differs from the sparse KP masks typically used in other tasks, such as human pose estimation. Sparse KP estimation involves assigning a probability density centered at each sparse KP location, achieved by applying a Gaussian kernel to the KP. Where each KP has a `class' and each class has a different output layer. Due to the unconstrained nature of chart KP, we output a single heatmap which makes it harder for the model to segment each point by only using sparse labels.

To address this issue, we create dense directional KP masks by interpolating KP between inflection points. Specifically, we interpolate 10 fixed KP between each ground truth point and use a Gaussian kernel with a spread of 2.0 to generate the dense mask. Fig. \ref{fig:mask} illustrates an example of the resulting dense mask.

To compute the loss during training, we use the dense masks to focus on informative pixels in the image. We binarize the masks to create classification labels for line plots using a threshold of 0.6. For bar plots, we only use the top-left, center, and bottom-right KP, and for box plots, we only use five KP for each marking. For scatter plots, we use all KP. To generate background masks, we invert the foreground masks.


\subsubsection{SpaDen Keypoint-Loss} 
\input{Sections/Figures/005_loss_heads}
utilizes a combination of direct regression, binary classification, and even multi-class classification. While each loss has its merits, each faces unique challenges on unconstrained KP when trained individually. The lack of data makes regression overfit. Multiple KP without class discriminator makes binary cross entropy hard to converge on a single heatmap, whereas a general lack of informative pixels $(99.9\%$ pixels in plot area are background) makes cross entropy an imbalanced classification problem. Even dice loss for tackling imbalanced classification severely affects the training process of the model when unconstrained KP from different plot area components are projected on the same single layer instead of the per-class output layer. 

Our final model consists of 5 heads for the different `views' of the chart. Direct binary reconstruction, fore/back-ground regression, and fore/back-ground classifier. 
Fig. \ref{fig:loss_he} shows output predictions for each loss individually as well as when combined. We weigh the background pixels to $0.01$ and the foreground to $0.99$. We find that using a mixture of sparse masks for classification, and dense directional masks for regression, works best. This architecture choice helps further with chart component reconstruction by providing specific feature outputs to calculate global chart vs. KP attention and also in post-processing for discrete and continuous data in the same model, as described in the next section.  
The final loss value is an alpha blend of $0.7\times$ Aggregated KP Loss from $5$ heads and $0.2\times$ contrastive loss for learning associations and $0.1\times$ chart type classification.  

\subsection{Keypoint Clustering} 
In this stage, we first post-process the output heatmaps and then use a per-pixel contrastive loss to cluster the KP.


\subsubsection{Foreground Keypoint Heatmaps} 
\input{Sections/Figures/007_Filter_regression} 
 are critical for our method. The regression output enables us to predict the complete contour of chart elements accurately. The classification output gives us corners/points for discrete elements. Foreground regression output is also used to extract clusters of relevant pixels of individual chart elements. To avoid noise interference in the clustering process, we apply a post-processing procedure shown in Fig. \ref{fig:filtereg}. Our approach involves identifying pixel clusters by selecting the top 1,000 pixels with the highest confidence values and reducing each connected component to individual points by thresholding at $0.85\times$ max-intensity. 
 
 We also find the median RGB color of the chart and discard all foreground KP within $0.25$ min-distance. For each point, we obtain the RGB color distribution and identify the median color value, as well as the peaks in the distribution except for the median. We discard all points with a color distance greater than the minimum distance between the median color and peaks. If the legend patch is available, we use the centroid distance from the median color to the mean legend patch color else mean peak value from cluster. Color distance is computed as the L2 distance between RGB tuples. 




\subsubsection{Chart Component Reconstruction} 
\input{Sections/Figures/006_ContrastiveKP}
is done from the pixel clusters learnt through contrastive loss. First, we calculate the cross-attention between the global chart features from the backbone and each of the individual heatmap heads $HM_i$. The resulting per-pixel embeddings are summed and aggregated to perform clustering, as shown in Fig. \ref{fig:contrKP}.

We treat all KP belonging to the same instance of a line, bar, box, or scatter as positive samples, and rest as negative. During inference, we filter the regression heatmap output to obtain $K$ informative pixels and then calculate their exhaustive $K \times K$ similarity. A threshold value of $0.85$ is used for cosine similarity in the MS loss or $1e^{-5}$ for Euclidean distance in the push-pull loss.

Each resulting cluster is reconstructed to its original chart component using heuristic rules. Lines are assumed to have a vertical axis as the dependent variable and are joined from left to right. For bars, boxes, and scatter plots, the closest point from each cluster with the highest probability output from the foreground classification heatmap is taken. For bars and boxes, we take the top 2 (corner) and top 5 (whiskers) points, respectively. For scatter plots, we threshold by selecting all points with a confidence interval of $0.25 \times$ the maximum value.

To retrieve the data, we use task-6 inputs of a text box, axis tick, and text role to get the data value at the pixel location. We also classify horizontal/vertical bars and boxes as such. 

\subsection{Legend Mapping} 
To effectively associate the chart components with their  legend patches, we introduce a ROI-align layer that operates on the backbone features. We utilize a legend oracle to obtain bounding box coordinates,  then the roi-align layer provides uniform embeddings as illustrated in Fig. \ref{fig:arch-llm}. We aggregate the KP clusters using concatenation and 1x1Conv-BN and use the MS-Loss function to measure their similarity with each of the legend patches. During training, the legend oracle provides the association label. At inference time, we  match the KP cluster with the most similar legend patch.

\subsection{Invariance to Chart Text} 

\input{Sections/Figures/003_dataaug.tex}
KP based backbones exhibit sensitivity to individual pixels, making invariance to text a desirable characteristic. To achieve this, the heatmap features for the plot area must generate KP exclusively for components within the chart area, and not for text within the chart. In order to evaluate the robustness of our approach, we introduced `easy' and `hard' samples by selectively adding or removing text boxes from the chart. During each iteration of the training process, there is a $25\%$ chance that all text boxes and content are replaced with the median color from the chart, and a $25\%$ chance of adding skewed and cropped contextual text boxes to random positions within the chart. The model may receive different augmented chart inputs in each epoch, but the same mask labels. The efficacy of this augmentation strategy is demonstrated in Figure \ref{fig:data_aug}, where the text box information is obtained from an oracle.

%% file: Sections/Figures/002_UDE.tex
\begin{figure}[ht!]
         \centering
         \includegraphics[width= 0.4\textwidth]{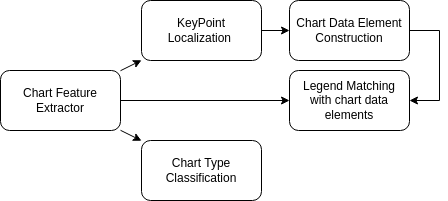}
         \caption{\scriptsize{UDE Framework Blocks}}
         \label{fig:ude}
\end{figure}



%% file: Sections/Figures/004_masks.tex

\begin{figure}[ht!]
         \centering
         \includegraphics[width= 0.9\textwidth]{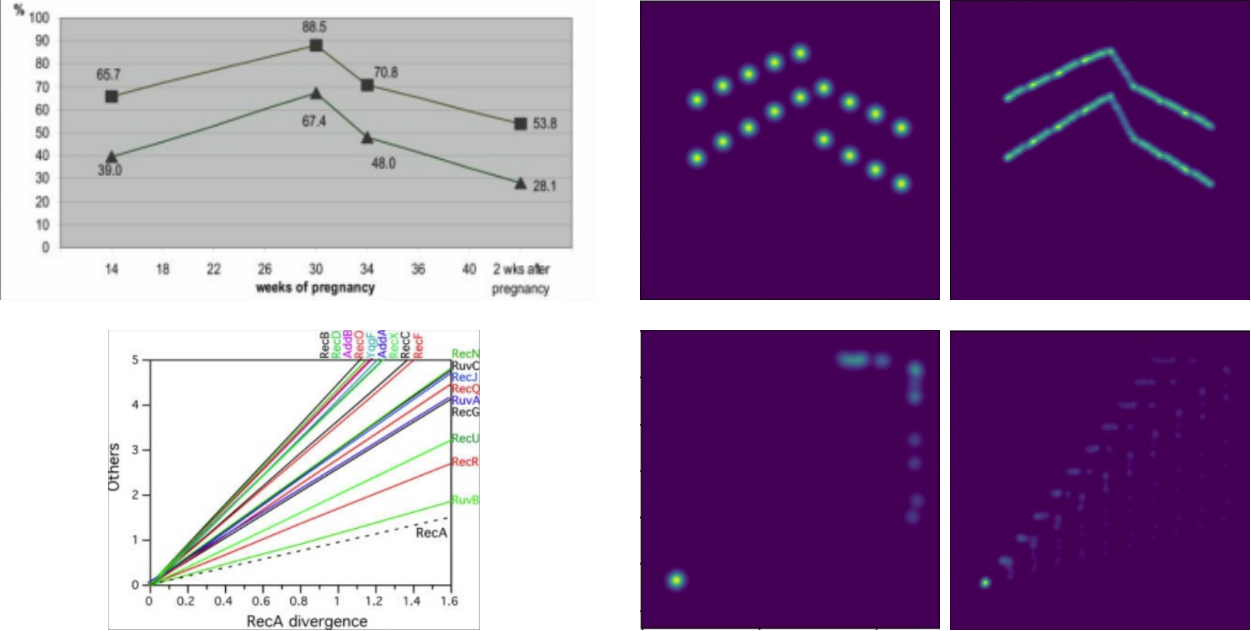}
         \caption{\scriptsize Custom Groundtruth Masks. First Column: Input Chart, Second Column: Generic Gaussian Mask(Classification), Third Column: Dense Directional Mask(Regression) .}
         \label{fig:mask}
\end{figure}

%% file: Sections/Figures/005_loss_heads.tex
\begin{figure}[ht!]
         \centering
         \includegraphics[width= \textwidth]{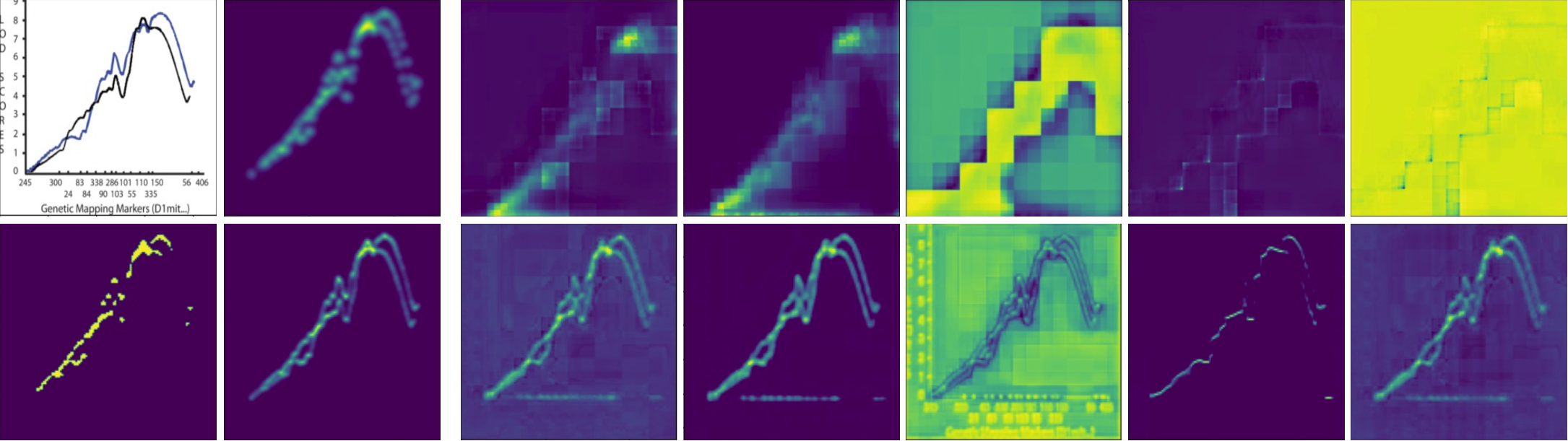}
         \caption{\scriptsize Affect of Keypoint loss. Top row shows output when trained in isolation : input chart, gaussian mask, predicted BCE heatmap, predicted MSE foreground, predicted MSE background, predicted CE foreground, predicted CE background. Bottom Row shows outputs when trained in combination, first two show the groundtruth Classification Mask and Regression Mask, rest are counterparts of top row outputs.}
         \label{fig:loss_he}
\end{figure}

%% file: Sections/Figures/007_Filter_regression.tex
\begin{figure}[ht!]
         \centering
         \includegraphics[width= 0.85\textwidth]{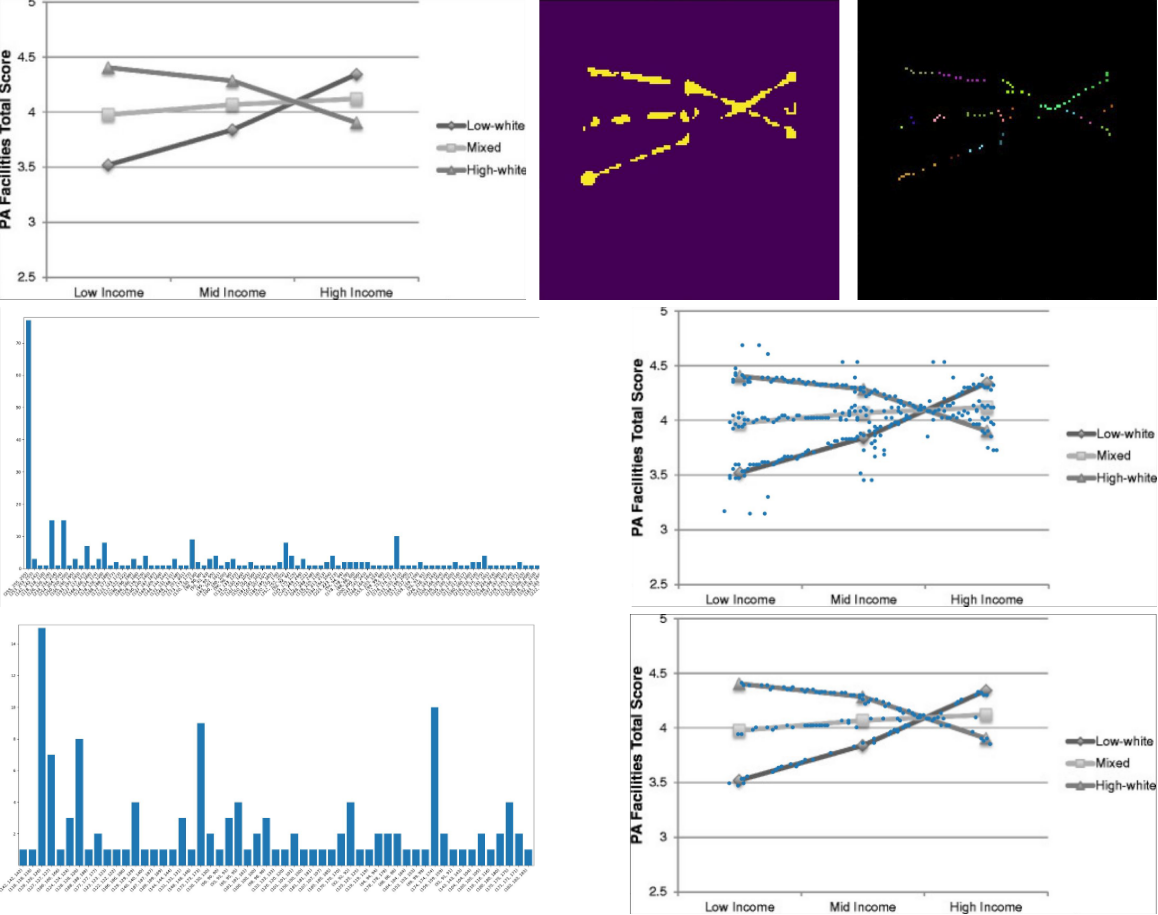}
         \caption{\scriptsize Post-Processing Regression Prediction : (Top row) Input image, Thresholded islands, Points from connected components, (Middle Row) Colour Histogram over all points, All points from cc before filtering, (Bottom Row) Histogram after discarding pixels close to median color, Final points chosen for contrastive grouping.}
         \label{fig:filtereg}
\end{figure}

%% file: Sections/Figures/006_ContrastiveKP.tex
\begin{figure}[ht!]
         \centering
         \includegraphics[width= 0.85\textwidth]{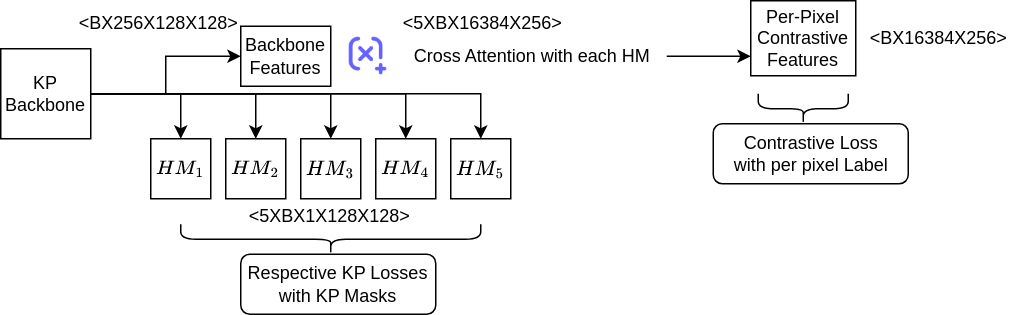}
         \caption{\scriptsize Global chart vs key-point attention implementation to generate Contrastive Feature Map }
         \label{fig:contrKP}
\end{figure}

%% file: Sections/Figures/003_dataaug.tex
\begin{figure}[ht!]
         \centering
         \includegraphics[width= 0.9\textwidth]{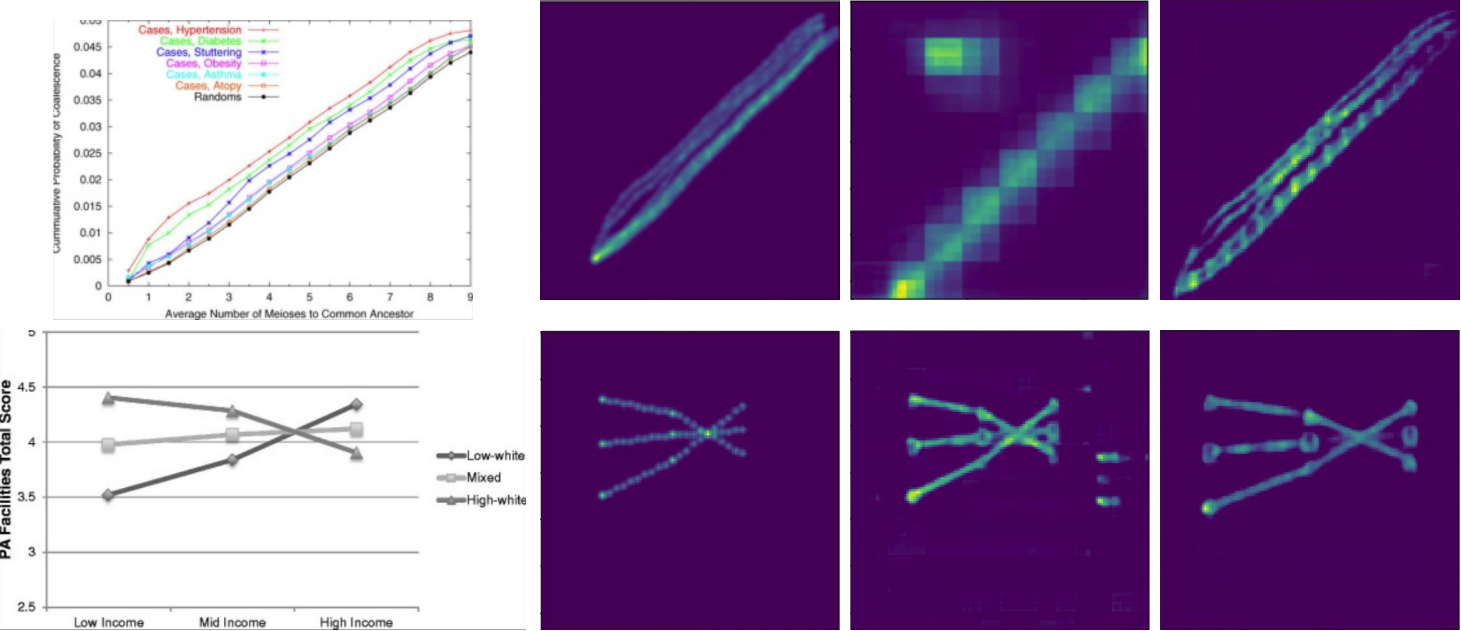}
         \caption{\scriptsize Text box invariance for chart Keypoint-Estimation. TopRow : Input Chart, Directional Mask, Predicted heatmap using generic mask w/o augmentation, Predicted heatmap using custom mask with augmentation. BottomRow : Input Chart, Directional Mask, Predicted heatmap using custom mask w/o augmentation, Predicted heatmap using custom mask with augmentation.}
         \label{fig:data_aug}
\end{figure}

%% file: Sections/004_Results.tex
\section{Experiments}
 We first describe the dataset and evaluation and then provide an exhaustive quantitative evaluation. We summarize our findings and conclusions comparing different architectures, pooling, aggregation techniques, objective functions, and post-processing methods for chart reconstruction.

\subsection{Dataset and Evaluation} 
\subsubsection{The Chart-Infographics Challenge Dataset} is a rigorously curated collection of over 86,815 real charts sourced from the Open Access (OA) section of PubMed Central (PMC). The dataset comprises both multi-panel and single-panel charts, with the latter forming the focus of downstream tasks. Among these, over 36,000 images are categorized as single-panel charts and are utilized for training and testing. Multi-panel charts are divided to extract under-represented chart classes. The final training set comprises 22,923 images, and the testing set comprises 13,260 images. In addition, we incorporate 999 charts from Ar$\chi$iv publications, as provided in \cite{siegel2016figureseer}, along with 100k synthetic charts from the previous Chart-Infographics '19 challenge, to enhance the training data. During each epoch, training is conducted over all real charts and an equivalent number of randomly selected synthetic charts.

\subsubsection{The Evaluation} is conducted on the publicly released test set \cite{icpr2022}, with the same splits as the challenge, ensuring a fair comparison in our study. The test set contains a diverse range of chart types, including line plots, scatter plots, horizontal and vertical bar charts, and vertical box-plots. The charts in the dataset have been annotated for hierarchical image classification and further categorized as containing charts or not. For the downstream tasks, the output and ground-truth for each chart are defined as the visual location in terms of $(x,y)$ coordinates for task-6a and a set of (name, data series) pairs for task-6b.

The evaluation metrics for the tasks are designed to capture the nuances of different chart types. For continuous data series, represented by line charts, the metric quantifies the difference between two functions as an integral of their point-wise differences. Bi-variate point set data, such as scatter plots, is evaluated using a capped distance function, where the distance is scaled by the inverse covariance matrix of the ground truth points. Discrete data, represented by bar charts and some line plots, is evaluated using two cases: exact text match and fuzzy text matching. The distance is calculated based on string equality for the exact match case and the normalized edit distance between predicted and ground truth strings for fuzzy text matching. Boxplots are evaluated using exact text matching between the predicted and ground truth strings for real-numbered summary statistics such as minimum, maximum, first quartile, third quartile, and median, ignoring outliers in this representation. The final metric is calculated as 1 minus the distance for each task.
 We refer the readers to the challenge for exhaustive details \footnote {\url{https://chartinfo.github.io/metrics/metric.pdf}} and publicly available script  
 \footnote {\url{https://github.com/chartinfo/chartinfo.github.io/blob/master/metrics/}} used for our evaluation.

\subsection{Experiment Result}

\subsubsection{Qualitative outputs} were shown in previous sections alongside the description of our implementation in Figs  \ref{fig:data_aug},  \ref{fig:loss_he},  \ref{fig:contrKP}. These are all conducted with the same simple HGN backbone without added pooling or aggregation. We further discuss quantitative metrics below.



\subsubsection{Exhaustive quantitative} evaluation is provided in Table \ref{tab01::main_res}, grouped by rows. 
\input{Sections/Tables/001.tex} We report most runs for the `Line' type chart as they are the hardest to reconstruct and further provide evaluation over all types for the best-performing combination for each flavor of backbone.

We evaluated different reconstruction strategies, ranging from heatmap output to chart objects, using various heuristics such as connected component analysis [CC], [HOG] features extracted from the original image after thresholding and clustering points, and low-level feature matching techniques using correlation, local binary patterns, and cross-correlation. Additionally, we tested push-pull vs. multi-similarity loss objectives for contrastive learning. 
All hyperparameters for backbone models is kept the same as the original implementations. All thresholding multipliers mentioned in section \ref{sec:UDE} are calculated using Otsu's hysteresis on randomly sampled heatmaps during validation, minimising val loss and maximising val accuracy. 

Rows $1-2$ benchmark the public cornernet (HGN+CP) implementation of \cite{luo2021chartocr}. Their performance on 6a is comparable out-of-the-box, but for 6b, they assume axis position and calculate the range of data values as the max difference between OCR outputs, which does not scale for real charts and has a low 6b-data score. The model also cannot perform legend matching and does not report a 6b-name score. We used an open-source implementation \cite{smith2007overview} to replace their proprietary text box localization and recognition software while keeping everything else the same.

Rows $3-6$ present partial results from the latest challenge participants, referred to as `IIT\_CVIT' in the challenge report\cite{icpr2022}.

Rows $7-42$ present our implementations of the proposed strategies described in this paper:

\begin{enumerate}
    \item Rows $7-8$: Baselines with simple ResNet32 and FCN based backbones.
    \item Rows $9-13$: We tested the best reconstruction strategy with a simple HGN backbone, and learned embedding through MS Loss provided the best results.
    \item Rows $14-22$: We added pooling and aggregation to the HGN backbone and compared across objective and chart types.
    \item Rows $23-31$: We compared CPN backbone-based models with contrastive loss, pooling, aggregation, and different chart types.
    \item Rows $32-42$: We conducted a similar study for SPN-based models. Since improving encoder architecture through pooling and aggregation did not help, we used a base model for different chart types.
\end{enumerate}

The results show that the best-performing model for line charts is the HGN + CP + DLA model with MS loss for both element detection and data extraction, achieving F1 scores of 0.83 and 0.69, respectively. 

The best-performing model for bar charts is also the HGN + CP + DLA model with the MS loss function, achieving an F1 score of 0.912 for element detection and 0.81 for data extraction.

For box charts, the best-performing model is the IIT\_CVIT model with heuristic-based element reconstruction, achieving an F1 score of 0.97 for element detection and 0.834 for data extraction. 

For scatter charts, the HGN + CP + DLA model with the MS loss function achieves the highest F1 score of 0.782 for element detection and 0.62 for data extraction.

The results also show that rule-based methods for element detection, such as connected components and histogram of gradients, perform poorly compared to learned methods. Furthermore, combining different models, such as the HGN + CP + DLA models, improves the overall performance of the system.
It is interesting to note that SPN (upsampling is learned through deconvolution) performs better without added pooling and aggregation. This shows that while these techniques help improve encoders, their contribution might be outperformed by a much more sophisticated decoder, which remains invariant to these features in the deconvolution operation.

\subsection{Conclusion}
We have described our approaches to chart data extraction through bottom-up KP parsing methods. We present an end-to-end framework for chart visual element detection, data series extraction, and legend matching. We have provided exhaustive experimentation with multiple backbones, pooling, and layering strategies.  We find that our approach using HG-Net KP backbone augmented with the proposed pooling and aggregation techniques performs the best. 

%% file: Sections/Tables/001.tex
\begin{table}[ht!]
\centering
\scriptsize
\caption{\scriptsize \textbf{Data Extraction Results} for different models on the chart element detection(6a) and data extraction task(6b). NN - Neural Net, CC - Connected Component, HOG - Histogram of Gradient, FM - Feature Matching, PP - Push Pull Loss, MS - Multi Similarity Loss }
\label{tab01::main_res}
\begin{tabular}{|c|c|c|c|c|c|c|}
\hline
\# & Model Backend  & Element Reconstruction & Chart Type & 6a & 6b-Data & 6b-Name\\ \hline

1 & Chart-OCR (pretrained)      & NN + Rules  & Line    & 0.71 & 0.25 & -\\ \hline 
2 & Chart-OCR (pretrained)      & NN + Rules  & Bar    & 0.88 & 0.28 & - \\ \hline  

3 & IIT\_CVIT (as reported) & Heuristic  & Line    & 0.773 & - & - \\ \hline
4 & IIT\_CVIT (as reported) & Heuristic  & Bar     & 0.906 & - & - \\ \hline
\rowcolor{gray!10}
5 & IIT\_CVIT (as reported) & Heuristic  & Box     & \textbf{0.970} & 0.834 & \textbf{0.921} \\ \hline
6 & IIT\_CVIT (as reported) & Heuristic  & Scatter & 0.773 & - & - \\ \hline
\hline

\hline
7 & ResNet32 & Rule based [CC] & Line    & 0.29  & 0.256 & 0.342 \\ \hline
8 & FCN & Rule based [CC] & Line    & 0.38 & 0.286 & 0.33 \\ \hline

9 & HGN & Rule based [CC]   & Line     & 0.49 & 0.294  & 0.34 \\ \hline
10 & HGN & Rule based [HOG] & Line    & 0.56 & 0.292 & 0.34\\\hline
11 & HGN & Rule based [FM]  & Line    & 0.14 & 0.285 & 0.34 \\ \hline
12 & HGN & Learnt [PP]      & Line    & 0.68 & 0.611 & 0.741 \\ \hline
13 & HGN & Learnt [MS]      & Line    & 0.699  & 0.683  & 0.756  \\ \hline
14 & HGN + CP & Rule based [CC] & Line &0.49  & 0.324 & 0.34 \\ \hline
15 & HGN + CP  & Learnt [PP]    & Line    &0.69  & 0.62 & 0.756 \\ \hline
16 & HGN + CP & Learnt [MS]     & Line    &0.71  & 0.685 & 0.77  \\ \hline
17 & HGN + CP + DLA & Learnt [PP]  & Line    & 0.74 & 0.66 & 0.75 \\ \hline
18 & HGN + CP + DLA  & Learnt [MS] & \textbf{Line}    &  \textbf{0.83} & \textbf{0.69} & \textbf{0.77} \\ \hline
19 & HGN + CP + DLA & Learnt [MS]       & \textbf{Bar}    & \textbf{0.912} & \textbf{0.81} & \textbf{0.86} \\ \hline
20 & HGN + CP + DLA & Learnt [MS]       & Box    &  0.965 & \textbf{0.882} & 0.88 \\ \hline
21 & HGN + CP + DLA & Learnt [MS]       & \textbf{Scatter}  & \textbf{0.782}  & \textbf{0.62} & \textbf{0.55} \\ \hline
22 & HGN + CP + DLA & Learnt [MS]       & \textbf{ALL}    & \textbf{0.8722} & \textbf{0.7505} & \textbf{0.765} \\ \hline
\hline

\hline
23 & CPN   & Learnt [PP]  & Line    & 0.67 & 0.581 & 0.701 \\ \hline
23 & CPN   & Learnt [MS]   & Line    & 0.66 & 0.592 & 0.69 \\ \hline
24 & CPN + CP  & Learnt [PP]  & Line & 0.67  &  0.585 & 0.701 \\ \hline
25 &CPN + CP  & Learnt [MS]  & Line   & 0.66  & 0.592 & 0.69 \\ \hline
26 &CPN + CP + DLA & Learnt [PP]       & Line    &  0.675 & 0.58  & 0.69 \\ \hline
27 &CPN + CP + DLA & Learnt [MS]       & Line    & 0.68 & 0.61 & 0.56 \\ \hline
28 &CPN + CP + DLA & Learnt [MS]       & Bar    & 0.89 & 0.78 & 0.72 \\ \hline
29 &CPN + CP + DLA & Learnt [MS]       & Box    &  0.952 & 0.83 & 0.772 \\ \hline
30 &CPN + CP + DLA & Learnt [MS]       & Scatter    & 0.71  & 0.601 & 0.661 \\ \hline
31 &CPN + CP + DLA & Learnt [MS]       & ALL    & 0.808 & 0.705 & 0.678  \\ \hline
\hline

\hline
32 & SPN    & Learnt [PP]       & Line    & 0.59 & 0.448 & 0.342 \\ \hline
33 &SPN    & Learnt [MS]       & Line    & 0.62 & 0.51 & 0.35 \\ \hline
34 &SPN + CP       & Learnt [PP]       & Line    & 0.587 & 0.44 & 0.342 \\ \hline
35 &SPN + CP       & Learnt [MS]       & Line   &0.613  & 0.505  &  0.32 \\ \hline
36 &SPN + CP + DLA   & Learnt [PP]   & Line  & 0.577 & 0.421 & 0.338 \\ \hline
37 &SPN + CP + DLA  & Learnt [MS]   & Line    & 0.601 & 0.499  & 0.336 \\ \hline
38 &SPN + CP + DLA    & Learnt [MS]  & ALL    & 0.623 & 0.59 & 0.482 \\ \hline

39 &SPN    & Learnt [MS]       & Bar    & 0.82  & 0.782 & 0.68 \\ \hline
40 &SPN    & Learnt [MS]       & Box    & 0.85 & 0.812 & 0.778 \\ \hline
41 &SPN    & Learnt [MS]       & Scatter    & 0.58 & 0.324 & 0.27 \\ \hline
42 &SPN    & Learnt [MS]       & All    & 0.7175 & 0.607 & 0.519 \\ \hline

\end{tabular}

\end{table}